\documentclass[journal, twoside, web]{ieeecolor}
\usepackage{generic}
\usepackage{xcolor}
\usepackage{cite}
\usepackage{subfiles}
\usepackage{svg}
\usepackage{url}
\usepackage{hyperref}
\usepackage{subfiles}
\usepackage{url}
\usepackage{enumerate}
\usepackage{subcaption}
\usepackage{amsmath, amssymb, amsfonts, mathtools, mathrsfs}
\usepackage{algorithm, algpseudocode}
\usepackage{bm, multirow, array}
\newtheorem{theorem}{Theorem}
\newtheorem{remark}{Remark}
\newtheorem{lemma}{Lemma}

\newtheorem{definition}{Definition}

\newtheorem{assumption}{Assumption}
\DeclareMathOperator*{\argmin}{arg\,min}
\newcommand{\norm}[1]{\lVert#1\rVert}
\def\BibTeX{{\rm B\kern-.05em{\sc i\kern-.025em b}\kern-.08em
    T\kern-.1667em\lower.7ex\hbox{E}\kern-.125emX}}
\begin{document}
\title{On Enhancing Structural Resilience of Multirobot Coverage Control with Bearing Rigidity}
\author{ 
Kartik A. Pant, Vishnu Vijay, Minhyun Cho, and Inseok Hwang
\thanks{The authors are with the School of Aeronautics and Astronautics, Purdue University,
West Lafayette, IN 47906. Email: ({\tt\small kpant}, {\tt\small vvijay}, {\tt\small cho515}, {\tt\small ihwang} {\tt \small@purdue.edu})}}
\maketitle
\thispagestyle{empty}
\begin{abstract}   
The problem of multi-robot coverage control has been widely studied to efficiently coordinate a team of robots to cover a desired area of interest. However, this problem faces significant challenges when some robots are lost or deviate from their desired formation during the mission due to faults or cyberattacks. Since a majority of multi-robot systems (MRSs) rely on communication and relative sensing for their efficient operation, a failure in one robot could result in a cascade of failures in the entire system. In this work, we propose a hierarchical framework for area coverage, combining centralized coordination by leveraging Voronoi partitioning with decentralized reference tracking model predictive control (MPC) for control design. In addition to reference tracking, the decentralized MPC also performs bearing maintenance to enforce a rigid MRS network, thereby enhancing the structural resilience, i.e., the ability to detect and mitigate the effects of localization errors and robot loss during the mission. Furthermore, we show that the resulting control architecture guarantees the recovery of the MRS network in the event of robot loss while maintaining a minimally rigid structure. The effectiveness of the proposed algorithm is validated through numerical simulations.     
\end{abstract}
\section{Introduction}
\label{sec:introduction}
Recent advances in multi-robot systems (MRSs), with their superior sensing, communication, and computational capabilities, allow them to perform complicated tasks otherwise impossible with only single-robot systems. MRSs have been widely adopted for numerous applications such as cooperative sensor coverage\cite{cortes2004coverage}, search and rescue\cite{kantor2006distributed}, and environmental monitoring \cite{burgard2005coordinated}. In recent catastrophic wildfires in Los Angeles, drone swarms have been actively utilized for monitoring and prevention of wildfires \cite{wildfire2025wong}. However, as the complexity of these systems increases, the number of failure modes affecting MRS performance and safety also increases. Furthermore, the sensing \cite{khan2023synthesis, pant2024adversarial}, and communication networks \cite{rezaee2024resilient} also open up new cyberattack surfaces, network vulnerabilities, and backdoors, which adversaries can exploit to degrade and disrupt the performance of the MRS. Thus, designing control architectures ensuring the system's resiliency under these unknown failure modes becomes essential. 


A key application of MRSs is to cover a desired area of interest, often denoted by a density function that indicates the importance of each point in the region. This task is termed as \textit{coverage control}. A group of robots deployed in an environment (in any arbitrary initial configuration) is moved to optimal locations that maximize the overall sensing/coverage performance.
Although extensive research on coverage control and its variants \cite{pierson2017adapting, schwager2009decentralized, todescato2017multi, lee2015multirobot, carron2020model, luo2019voronoi} have been done in the literature, the results often assume that all the robots remain healthy, connected, and free of localization errors throughout the operation, which may not be applicable in real-world situations. In many cases, the MRSs can fail due to known/unknown failure modes in navigation, communication, or control. For example, suppose the position estimates of some of the UAVs in a swarm are compromised by entering a GNSS-denied region or by an adversary injecting GNSS spoofing attack. One way to mitigate such failures is to take advantage of the inherent redundancy, i.e., locally sensed inter-robot measurements that exist by virtue of the configuration of robots \cite{vijay2025range}. To effectively utilize this information redundancy, it becomes essential for the network to maintain a resilient structure (or rigid structure), one where robot loss or anomalies can be easily detected and mitigated. 

\begin{figure}[t]
\centering
\includegraphics[width=0.5\textwidth,clip]{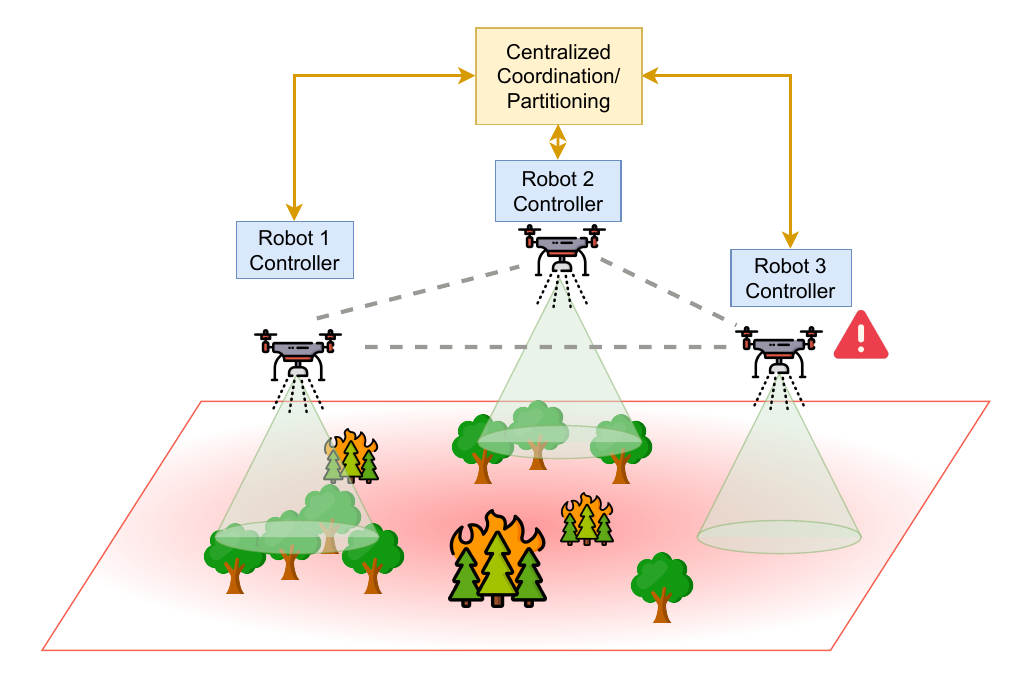}
\caption{Illustration of multi-robot coverage control for environmental monitoring during wildfire. Our proposed distributed approach ensures the coordination of the network of robots even when some robots become unhealthy while executing the mission. }
\label{fig:illustration}
\end{figure} 

In classical works, the coverage control has been rigorously studied with optimal solutions characterized by Centroidal Voronoi tesselation (CVT) \cite{du2006convergence}. In \cite{cortes2004coverage}, the authors proposed an iterative gradient-based control law using Lloyd's algorithm \cite{lloyd1982least} for the single integrator dynamics to converge to the Voronoi centroids. Coverage control with time-varying sensory function is analyzed in \cite{lee2015multirobot}, and a deep learning-based approach to cover an area with an unknown density function is proposed in \cite{rastgoftar2025deep}. The authors in \cite{schwager2009decentralized,todescato2017multi,rickenbach2024active} designed coverage algorithms that simultaneously estimate the sensory function along with the control design. Obstacle avoidance in coverage control is addressed in \cite{bai2021adaptive, bai2023safe}, where the authors designed a safe control strategy leveraging control barrier functions (CBFs). In \cite{rezaee2024resilient}, a distributed resilient strategy is proposed to mitigate the impact of misinformation in the network caused by cyberattacks on a subset of robots. Recently, the authors in \cite{carron2020model, rickenbach2024active} considered the coverage problem with the nonlinear robot dynamics with state and input constraints and formulated it as a reference tracking model predictive control (MPC). In most of the above approaches, the underlying network topology is implicitly assumed, and the robots stay healthy throughout the coverage task.

Significant gaps exist in the literature where coverage control is studied while maintaining a resilient network topology. In \cite{luo2019voronoi}, a connectivity-constrained coverage controller is proposed that enforces connectivity by imposing hard constraints as barrier certificates using the algebraic network connectivity, i.e., eigenvalues of the graph Laplacians. However, such an approach can lead to infeasibility and deadlocks \cite{cavorsi2023multi}. To address this issue, the authors in \cite{cavorsi2023multi} utilized a bypass mechanism that considers an alternate CBF to relax connectivity constraints, minimizing deadlocks. However, it requires an exact representation of the environment, which may not be suitable for coverage control.
In this work, we focus on the aforementioned challenges and tackle the problem of enhancing the structural resilience of the underlying network topology of the MRS in a coverage control setting. Our goal is to move the robots to cover a desired area while ensuring that they remain well-connected and collaboratively protect/guide each other under unknown localization drifts/failures caused by faults or cyberattacks. We transform this problem into rigidity maintenance of a network (or structure) formed by robots (as joints) and their virtual sensing and communication links (as bars) as it is scaled and translated in space while viewing the localization failures and robot loss as perturbation forces acting on it.

We take advantage of the rigidity theory originally developed to analyze the flexibility (deformability) of rigid structures arising from external perturbations. Rigidity theory has been previously investigated in the MRS literature, notably for decentralized self-localization\cite{zhao2015control}, formation stabilization \cite{zhao2015bearing,schiano2016rigidity}, integrity monitoring \cite{vijay2025range} and reconfiguration \cite{anderson2008rigid}. In this work, for the first time, we leverage concepts from rigidity theory for coverage control. We impose the MRS's motion to satisfy the constant relative bearing between the robot's neighbors, i.e., bearing rigidity, which allows the network to scale and translate to achieve desired coverage performance.
As a consequence of maintaining the rigid structure, we show that the MRS can swiftly reconfigure itself when a robot leaves the network by making new local connections (among two-hop neighbors) while preserving the structural/topological network properties, i.e., resiliency and connectivity.

Our main contributions can be summarized as follows: 
\begin{enumerate}
    \item We propose a distributed coverage control algorithm utilizing nonlinear tracking MPC that converges to the centroidal Voronoi configuration while constantly maintaining a resilient bearing-rigid network topology.
    \item We prove the convergence of the proposed distributed approach to the CVT (if feasible), otherwise to the nearest position to the CVT satisfying state and input constraints. Furthermore, we provide a design procedure to compute the terminal ingredients necessary to ensure the recursive feasibility and closed-loop stability of the distributed nonlinear tracking MPC.  
    \item We design a proactive rigidity recovery algorithm that iteratively finds the set of new recovery edges for each robot in the event of losing its neighbor due to faults or cyberattacks. 
\end{enumerate}

The rest of this paper is organized as follows. Section \ref{sec:prelim} briefly introduces the concepts from bearing rigidity theory. Section \ref{sec:prob_form} presents the problem formulation for multi-robot resilient coverage control. Section \ref{sec:resilient_mpc} provides the design of our proposed control architecture that enforces a bearing-rigid formation for the coverage task to improve the resilience of the underlying network topology. We also present a rigidity recovery algorithm that equips each robot with a set of new edges (connections) to be made in the case of a robot failure. Section \ref{sec:sim} provides an illustrative numerical example of our proposed approach. Section \ref{sec:conc} concludes the paper.

\textit{Notations:} 
Let $\mathcal{G} = \{ \mathcal{V}, \mathcal{E}\}$ denote an undirected graph with $\mathcal{V}$ as its vertices and $\mathcal{E} \subseteq \mathcal{V}\times\mathcal{V}$ as its edge set with $n = |\mathcal{V}|$ and $m = |\mathcal{E}|$. The set of neighbours of a vertex $i$ is denoted as $\mathcal{N}_i \triangleq \{j\in \mathcal{V}:(i,j)\in \mathcal{E} \}$. For any real-valued vector $x \in \mathbb{R}^n$  and positive definite matrix W, $\norm{x}_{W}^2 := x^\top Wx$ and $\norm{ \cdot }$ denotes the Euclidean norm. $\mathbb{R}_{+}$ denotes the set of positive real numbers. A class $\mathcal{K}$ function $\alpha: \mathbb{R}_+ \rightarrow \mathbb{R}_+$ is a continuous and strictly increasing function with $\alpha(0) = 0$. A class $\mathcal{K}_\infty$ function $\beta: \mathbb{R}_+ \rightarrow \mathbb{R}_+$ is a class $\mathcal{K}$ function and is unbounded.
\section{Preliminaries}
\label{sec:prelim}
In this section, we present important notations and results from bearing rigidity theory \cite{zhao2015bearing, trinh2019minimal}. For further details, we refer the reader to \cite{michieletto2021unified}.
\subsection{Bearing Rigidity Theory}
Consider a finite collection of $n$ points in $\mathbb{R}^d$ $(n\geq2, d\geq2)$ denoted as $\{p_i\}_{i=1}^n$. A configuration is defined as $\mathbf{p} = [p_1^\top, \dots, p_n^\top]^\top \in \mathbb{R}^{dn}$. Using a configuration $\mathbf{p}$ and an undirected graph $\mathcal{G}$, we define a framework as  
$\mathcal{G}(\mathbf{p})$, where each of vertex $i$ is uniquely mapped to each of the positions $p_i$, respectively.
Intuitively, a framework combines an abstract graph structure with a topology defined by a configuration. Given a framework $\mathcal{G}(\mathbf{p})$, we define
\begin{equation}
    \label{eq:bearing}
    e_{ij}:=p_j - p_i, \quad g_{ij}:= \frac{e_{ij}}{\norm{e_{ij}}}, \quad \forall (i,j) \in \mathcal{E},
\end{equation}
where the unit vector $g_{ij}$ represents the relative bearing of $p_j$ to $p_i$. Note that $e_{ij} = -e_{ji}$ and $g_{ij} = - g_{ji}$. 
Define the bearing function $f_{B}: \mathbb{R}^{dn} \rightarrow \mathbb{R}^{dm}$ as: 
\begin{equation}
    \label{eq:bearing_func}
    f_B(\mathbf{p}) = [g_1^\top, \dots, g_m^\top]^\top,
\end{equation}
where $m=|\mathcal{E}|$ is the number of edges in the framework and $\mathbf{p} = [p_1^\top, \dots, p_n^\top]^\top \in \mathbb{R}^{dn}$.
The bearing function \eqref{eq:bearing_func} describes the bearing of the entire framework. The following Jacobian is defined as the \textit{bearing rigidity matrix}: 
\begin{equation}
    \label{eq:bearing_matrix}
    R_B(\mathbf{p}):= \frac{\partial f_B(\mathbf{p})}{\partial \mathbf{p}} \in \mathbb{R}^{dm\times dn}.
\end{equation}

With the above notations defined, we now present some important properties and results of the bearing rigidity matrix.
\begin{lemma}[Bearing Rigidity Criterion\cite{zhao2015bearing}]
For a given framework defined by $\mathcal{G}(\mathbf{p})$, the bearing rigidity matrix satisfies the following rank criterion: $\text{rank}(R_B) \leq dn-d-1$.
\end{lemma}

We now formally present the definition of infinitesimal bearing rigidity and minimal bearing rigidity.
\begin{definition}[Infinitesimal Bearing Rigidity \cite{zhao2015bearing}]
We call a framework infinitesimally bearing rigid if its bearing motions are trivial, i.e., translation and scaling.
\end{definition}

The following definitions will introduce the concept of minimal bearing rigid graphs, which will be subsequently used to design network topologies with a minimum number of communication and sensing links composing a rigid structure. First, we define a generically bearing rigid graph. 
\begin{definition}[Generically Bearing Rigid Graph \cite{trinh2019minimal}]
A generically bearing rigid graph $\mathcal{G}$ is one that allows at least one configuration $\mathbf{p}\in \mathbb{R}^{dn}$ such that the framework $\mathcal{G}(\mathbf{p})$ is infinitesimally bearing rigid.
\end{definition}
\begin{definition}[Minimal Bearing Rigid Graph \cite{trinh2019minimal}]
A graph $\mathcal{G}$ composed of $n$ vertices and $m$ edges is minimally bearing rigid if and only if there does not exist any generically bearing rigid graph $\mathcal{H}$ in $\mathbb{R}^d$ with $n$ vertices and contains less than $m$ edges.   
\end{definition}
\subsection{Laman Graph and Henneberg Construction}
\label{subsec:henneberg}
We now review the iterative procedure to construct minimally bearing rigid graphs. First, we provide a combinatorial definition of Laman graphs, which represents a class of minimally rigid graphs.
\begin{definition}[Laman Graph \cite{zhao2017laman}]
A graph $\mathcal{G} = (\mathcal{V}, \mathcal{E})$ is a Laman graph if for all $k$, every subset of $k\geq2$ vertices spans at most $2k-3$ edges and the graph satisfies $|\mathcal{E}| = 2|\mathcal{V}| - 3$.
\end{definition}

\begin{figure}[h]
\centering
\includegraphics[width=0.5\textwidth,clip]{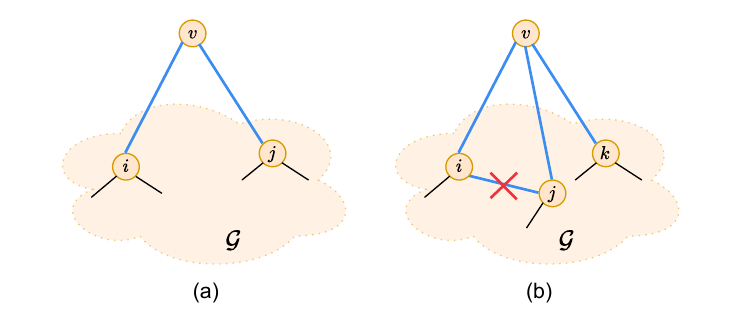}
\caption{Illustration of Henneberg construction to construct minimally rigid graphs utilizing two operations: (a) vertex-addition and (b) edge-splitting.}
\label{fig:henneberg}
\end{figure}

Intuitively, it implies that the graph's edges must be evenly distributed in a Laman graph. Another way to characterize Laman graphs is through Henneberg's construction. Figure \ref{fig:henneberg} illustrates the two operations of the Henneberg construction. 
\begin{definition}[Henneberg Construction \cite{zhao2017laman}]
For a given graph $\mathcal{G} = (\mathcal{V}, \mathcal{E})$, we form a new graph $\mathcal{G^{'}} = (\mathcal{V}^{'}, \mathcal{E}^{'})$ by appending a new vertex $v$ to the original graph $\mathcal{G}$ utilizing one of the following two operations:
\begin{enumerate}[(a)]
    \item Vertex Addition: Pick two vertices $i,j\in\mathcal{V}$ from graph $\mathcal{G}$ and connect it to $v$. This will lead to the resulting graph $\mathcal{G}^{'}$ with $\mathcal{V}^{'}= \mathcal{V} \cup v$ and $\mathcal{E}^{'} = \mathcal{E}\cup\{(v, i),(v, j)\}$. 
    \item Edge Splitting: Pick three vertices $i,j, k\in\mathcal{V}$ and an edge $(i,j) \in \mathcal{E}$ from graph $\mathcal{G}$. Connect each vertex to $v$ and delete the edge $(i,j)$. In this case, the resulting graph $\mathcal{G}^{'}$ will have $\mathcal{V}^{'} = \mathcal{V} \cup v$ and $\mathcal{E}^{'} = \mathcal{E}\cup\{(v, i),(v, j), (v, k)\} \backslash \{ (i,j)\}$. 
\end{enumerate}
\end{definition}
The resulting graph obtained by applying Henneberg construction, starting with an edge connecting two vertices, generates a Laman graph, and its converse is also true \cite{zhao2017laman}.
\section{Problem Formulation}
\label{sec:prob_form}
In this section, we present the problem formulation for resilient multi-robot coverage control. 
We consider a multi-robot system with $n$ robots that can move in the region described by a polytope $\mathcal{Q}\subset \mathbb{R}^d$, with $d =2$ or $d = 3$. Each robot in the MRS is governed by the discrete-time nonlinear dynamics. The $i^\text{th}$ robot's dynamics is given by:
\begin{align} \label{eq:agent_dynamics}
    x_{i,k+1} &= f_{i}(x_{i,k}, u_{i,k})\\
     p_{i,k} &= C_{i}x_{i,k}
\end{align}
where  $x_{i,k} \in \mathbb{R}^{n_x}$, $u_{i,k} \in \mathbb{R}^{n_u}$ and $p_{i,k} \in \mathbb{R}^d$ are the state, the control input and the position of robot $i$, respectively. The matrix $C_i \in \mathbb{R}^{d \times n_x}$ picks the position from the state. All robots are subject to state and input constraints, i.e., $(x_{i,k},u_{i,k})  \in \mathcal{Z}_i$ for all $k\geq 0$, where $\mathcal{Z}_i \subset \mathbb{R}^{n_{x}+n_{u}}$ is a compact set.

The steady-state values of the state, input and reference position of the robots $\bar{x}_{i}, \bar{u}_{i}$ and $\bar{r}_{i}$ are defined such that the robot dynamics \eqref{eq:agent_dynamics} is satisfied, i.e.,
\begin{align} \label{eq:agent_dynamics_ss}
    \bar{x}_{i} &= f_{i}(\bar{x}_{i}, \bar{u}_{i})\\
     \bar{r}_{i} &= C_{i}\bar{x}_{i}
\end{align}

\begin{assumption}
\label{ass:invariant}
The dynamics of each robot $i$ \eqref{eq:agent_dynamics} is position invariant, that is, $x_{i,k+1} + \psi(p_{i,k}) = f_i(x_{i,k} + \psi(p_{i,k}),u_{i,k}) \ \forall p_{i,k} \in \mathbb{R}^d, x_{i,k}, u_{i,k}$, where $\psi(p): \mathbb{R}^d \rightarrow\mathbb{R}^{n_{x}}$ represents an operator that takes the position of robots as input and generates a vector of the dimension of the state of the robots, $\psi(p_{i,k}) = [p_{i,k}^\top, 0, \dots,0]^\top \in \mathbb{R}^{n_x}$.
\end{assumption}

As a consequence of Assumption \ref{ass:invariant}, the robots are restricted to have the dynamics where the position appears only in the integrator. In practice, most robotic systems, such as rovers, aerial vehicles, and underwater robots, satisfy this assumption.
\begin{assumption}
\label{ass:dynamics}
The robot's dynamics $f_i$ are Lipschitz continuous with Lipschitz constant $\mathcal{L}_i$. Furthermore, it is assumed that each robot's state can be measured perfectly.
\end{assumption}

For the sake of simplicity, we consider the MRS's sensing and communication network to be identical and described by a minimally rigid graph $\mathcal{G} = (\mathcal{V},\mathcal{E})$. The graph $\mathcal{G}$ is constructed using the Henneberg construction defined earlier in Section \ref{subsec:henneberg}. With the graph vertices representing the robots of the multi-robot system, the edges represent the measurements that can be computed, as well as the communication links between the robots. 
\subsection{Voronoi-based Coverage Control}
Let the robot positions at time $k$ be denoted as $\mathbf{p}_k = [p_{1,k}^\top, \dots, p_{n,k}^\top]^\top \in \mathbb{R}^{n_x\times d}$. Let $\mathcal{O}_k = \{\mathcal{O}_{1,k}, \dots, \mathcal{O}_{n,k} \}$ be an arbitrary collection of polytopes that partition $\mathcal{Q}$. The coverage control problem can then be formalized as follows:
\begin{equation}
\label{eq:coverage_prob}
    \min_{\mathbf{p}_k,\mathcal{O}_k} \sum_{i=1}^n \int_{\mathcal{O}_{i,k}} \underbrace{\lambda(\norm{q - p_{i,k}}) \phi(q)dq}_{H(\mathbf{p}_k,\mathcal{O}_k)}, \quad p_{i,k} \in \mathcal{O}_{i,k},
\end{equation}
where $\phi: \mathcal{Q} \rightarrow\mathbb{R}_+$ is the \textit{density function} (or the information function) on the desired area $\mathcal{Q}$, $\lambda(\cdot)$ denotes the coverage performance of the robot at point $q$ from $p_{i,k}$, and $H(\mathbf{p}_k,\mathcal{O}_k)$ denotes the \textit{locational optimization cost}. Note that the optimal set $\mathcal{O}$ in \eqref{eq:coverage_prob}, when $\lambda(\cdot)$ is considered as the squared Euclidean norm, is given by the Voronoi tessellation \cite{cortes2004coverage,du2006convergence}. For a given position configuration $\mathbf{p}_k$, the partitions are denoted by $\mathcal{W}(\mathbf{p}_k) = \{ \mathcal{W}_{1}(p_{1,k}), \dots, \mathcal{W}_{n}(p_{n,k})\}$, with
\begin{equation}
    \mathcal{W}_{i}(\mathbf{p}_k) = \{ q \in \mathcal{Q} \ | \ \norm{q-p_{i,k}} \leq \norm{q - p_{j,k}}, \forall j  \neq i\}, 
\end{equation}
where $\norm{\cdot}$ denotes the Euclidean norm. If the configuration space $\mathcal{Q}$ is convex, all the sets in the partition $\mathcal{W}(\mathbf{p}_k)$ are also convex \cite{du2006convergence}. The optimal position of each robot is given by the centroids defined by $\mathbf{c}(\mathcal{W}(\mathbf{p}_k), \phi) = \big[c_1(\mathcal{W}_1(p_{1,k}), \dots, c_M(\mathcal{W}_n(p_{n,k})\big]$ with 
\begin{equation}
\label{eq:centroid}
    c_i(\mathcal{W}_i(\mathbf{p}_k), \phi) = \left( \int_{\mathcal{W}_i(\mathbf{p}_k)} \phi(q)dq\right)^{-1} \left(\int_{\mathcal{W}_i(\mathbf{p}_k)}q\phi(q)dq \right).
\end{equation}
The resulting locally optimal solution utilizing Voronoi tesselation for both partitioning and positions of the robots is called a \textit{centroidal Voronoi configuration}. Thus, the solution to the coverage problem defined in \eqref{eq:coverage_prob} reduces to ensuring that the position of each robot $p_i$ converges to the centroid of the Voronoi configuration $\mathbf{c}(\mathcal{W}_i(p_{i,k}), \phi)$. For the single integrator dynamics, this is achieved using Lloyd's algorithm \cite{cortes2004coverage}, which iteratively updates the position of each of the robots as $p_{i,k+1} = c_i(\mathcal{W}_i(p_{i,k}), \phi)$. For the nonlinear dynamics with input and state constraints, the convergence to the centroid can be obtained by utilizing a nonlinear output-tracking MPC controller \cite{carron2020model}. 

In most of the existing literature, the underlying network in a coverage control setting is assumed to be connected, and all robots are considered healthy, i.e., exhibit no failure/anomaly, which may not be practical. To this end, we focus on enhancing the resilience of an MRS network for coverage control tasks, particularly with minimal communication and sensing constraints, allowing robots to have a greater degree of freedom. The main challenge is to design a distributed algorithm that not only ensures optimal coverage but also ensures the robot stays connected and guarantees the system's recovery in case of a robot loss. This will allow the remaining robots to reorganize and accomplish the coverage task despite unknown failures on a subset of robots. Thus, the main goal is to design a distributed and resilient coverage control algorithm for the nonlinear robot dynamics, which ensures the optimality of coverage, connectivity, and self-healing capabilities in the event of robot failures. 
\begin{remark}
Note that in this work, we focus on a single robot loss at any given time and devise a recovery mechanism that ensures guaranteed network reconfiguration with minimal additional links. We posit that designing decentralized recovery mechanisms for simultaneous multiple robot losses is a significantly challenging problem, as it can result in highly abrupt changes in the network topology. We defer it for our future work.
\end{remark}

\section{Resilient Multi-robot Coverage Control}
\label{sec:resilient_mpc}
In this section, we will present our proposed control architecture and show that the resulting controller guarantees the convergence of the robot positions to the optimal Voronoi configuration. Our approach leverages nonlinear tracking MPC control for coverage control. Note that, we introduce an additional bearing maintenance cost in the optimization process to ensure that the robots maintain a bearing-rigid network topology. This ensures that even if a subset of robots becomes faulty or are lost during the mission, the rest of healthy robots can help mitigate localization failures or even reconfigure the MRS network (if the robots are lost) so that the desired coverage performance is maintained. This transforms the overall problem into a sequential multi-objective optimization problem, i.e., jointly minimizing the tracking and inter-robot bearing errors. 
\subsection{Nonlinear Tracking MPC with Bearing Maintenance}
We utilize the nonlinear tracking MPC framework to track the centroids defined in \eqref{eq:centroid} while maintaining a constant bearing. The centroids are updated iteratively based on a partition update criterion described later, ensuring robots converge to the optimal Voronoi partitions while satisfying recursive feasibility and closed-loop stability. 

The partition update step requires centralized coordination as it involves the knowledge of all robot positions to compute the Voronoi centroids. 
We define the computed reference setpoint as $r_{i,k} \coloneq c_i \big( \mathcal{W}(\mathbf{p}_{k}) \big)$, where $\mathbf{p}_k$ denotes the robot positions at time $k$. Let the tracking error between the robot positions and the computed centroids at time $k$ be defined as
\begin{equation}
    e_{i,k} = \norm{p_{i,k} - r_{i,k}}.
\end{equation}
To guarantee convergence, we utilize the partition criterion developed in \cite[Sec. 4]{carron2020model}. The partition is updated only if,
\begin{itemize}
    \item$\norm{p_{i,k} - r_{i,k}} \leq   e_{i,k} \quad \forall i \in \{1,\dots,n\},$  
    \item $\exists j \in \{1,\dots,n\} \quad \text{s.t.} \  \norm{p_{j,k} - r_{i,k}} \leq   e_{j,k}.$
\end{itemize}
For bearing maintenance, we introduce an additional cost in the overall optimization. The computed Voronoi centroids \eqref{eq:centroid} are utilized to establish a bearing-rigid formation, which is then used to calculate the desired bearing vector. This can be described as follows. Consider a bearing rigid framework $\mathcal{G}(\mathbf{p}_k)$, defined by the robot's positions $\mathbf{p}_k$ at the time $k$ and the communication/sensing graph $\mathcal{G}$. Let us denote the desired minimally rigid framework as $\mathcal{G}(\mathbf{r}_{k})$, where $\mathbf{r}_{k}$ is the Voronoi centroids after the partition update. Thus, the desired bearing of robot $i$ is denoted as $g_{i,k}$, which is obtained using the bearing function $g_{i,k} = f_B(\mathbf{r}_{k})$.

In this work, we formulate the resilient coverage control strategy for simultaneous centroid tracking and bearing maintenance as a nonlinear output tracking MPC problem. We define the artificial steady-state state-input pair for each robot as $(\bar{x}_i, \bar{u}_i)$. The steady-state position of each robot is $\bar{r}_i$. Given a position reference $r_{i,k}$, which is set as the centroids of the partition for each robot and a bearing reference $g_{i,k}$ is computed using $r_{i,k}$, the optimization problem computes the control input $u_{i,k}$ along with the steady-state state-input pair $(\bar{x}_i, \bar{u}_i)$ and an artificial reference $\bar{r}_i$ that tracks $r_{i,k}$. 
\begin{remark}
By choosing the position reference for Voronoi centroids $\bar{r}_i$ as an additional decision variable in the optimal control problem, it is ensured that any changes in the centroids resulting from the partition update over time do not affect the feasibility of the optimization.
\end{remark}

To simplify the notations, we define $J_i(\cdot) \coloneq J_i(x_{i,k}, r_{i,k}, g_{i,k}; \mathbf{u}_i, \bar{r}_i)$, where $x_{i,k}, \ r_{i,k}$, and $g_{i,k}$ are inputs, and $\mathbf{u}_i$ and $\bar{r}_i$ are outputs. Let $N$ be the prediction horizon for the MPC. We denote $x_{i,l|k}$ and $u_{i,l|k}$ as the $l^{\text{th}}$ step forward prediction of the state and inputs of the robot $i$ at time $k$, respectively, with $l \in \{1,\dots, N\}$, where $x_{i,0|k} = x_{i,k}$ and $\mathbf{u}_i = [u_{i,0|k}^\top, u_{i,1|k}^\top, \dots, u_{i,N|k}^\top]^\top$. Then, the overall cost function for each robot can defined as:
\begin{align}
\label{eq:mpc_cost}
 J_i(\cdot) = \sum_{l=0}^{N-1} &\ell_{i,l}(x_{i,l|k} - \bar{x}_{i}, u_{i,l|k} -  \bar{u}_i) + \ell_{i,N}(x_{i,l|k} - \bar{x}_i, \bar{r}_i) \nonumber
 \\+ \mu &\ell_{i,r}(r_{i,k} - \bar{r}_i) + (1-\mu)\ell_{i,b}(g_{i,k}, \bar{r}_i),    
\end{align}
where $\ell_{i,l}: \mathbb{R}^{n_x} \times \mathbb{R}^{n_u} \rightarrow \mathbb{R}$ is the stage cost, $\ell_{i,N}: \mathbb{R}^{n_x} \times \mathbb{R}^{d} \rightarrow \mathbb{R}$ is the terminal cost, $\ell_{i,r}: \mathbb{R}^{d} \rightarrow \mathbb{R}$ is the reference centroid tracking cost, and $\ell_{i,b}: \mathbb{R}^{dn_x} \times \mathbb{R}^{d} \rightarrow \mathbb{R}$ is the bearing maintenance cost. All of the above are positive definite functions. $\mu\in(0,1]$ denotes a constant, which ensures that the combined cost is a convex combination of the reference tracking and bearing maintenance costs. Note that the first two terms of $J_i(\cdot)$ penalize the deviation with respect to the artificial position reference $\bar{r}_i$. In contrast, $\ell_{i,r}(\cdot)$ penalizes the deviation between the true centroid and the artificial reference $(r_{i,k} - \bar{r}_i)$. In our formulation, the bearing maintenance cost depends on the equilibrium reference $\bar{r}_i$ instead of introducing an additional equilibrium bearing output vector as a decision variable. This crucial step allows us to construct a coupling between the centroid reference tracking and bearing maintenance costs. Moreover, keeping $\mu>0$ ensures that the reference centroid tracking is always achieved. Furthermore, by constructing a convex combination of the two costs, the overall cost is convex and has a unique minimizer, i.e., one that ensures that the robots reach the centroidal Voronoi configuration.

We introduce the following restricted sets essential for deriving the convergence result of our proposed approach. In order to avoid the above steady-state variables for each robot $i$ with active constraints, we define the following restricted set
\begin{equation}
    \hat{\mathcal{Z}_i} = \{ z \ | \ z + e \in \mathcal{Z}_i, \forall \norm{e} \leq \epsilon \}, 
\end{equation}
where $\epsilon>0$ is a small positive constant. We define $\bar{\mathcal{Z}}_i$ as the set of steady-state state-input pairs satisfying constraints, and $\bar{\mathcal{R}}_i$ as the set of corresponding steady-state positions where the state and input constraints are not active as follows:
\begin{align}
    \bar{\mathcal{Z}}_i &= \{z \ | \ z = (x,u) \in \hat{\mathcal{Z}_i}, \forall \norm{e} \leq \epsilon \}, \\
    \bar{\mathcal{R}}_i &= \{p \ | \ p = C_ix, (x,u) \in \bar{\mathcal{Z}}_i\}.
\end{align}

Note that since $\epsilon$ can be arbitrarily small, $\bar{\mathcal{Z}}_i$ will contain all steady-state state-input pairs close to the boundary of hard constraints defined by the set $\mathcal{Z}_i$. These sets will be utilized later as we show the convergence of our proposed approach.

To derive the stability of the proposed controller, we define an invariant set for centroid tracking by extending the idea of an invariant output tracking set introduced in \cite{limon2018nonlinear, limon2008mpc}.
\begin{definition}[Invariant Set for Centroid Tracking]
Let each robot $i\in \{1,\dots, n\}$ in the MRS have the constraint set on the state and the input as $\mathcal{Z}$, a set of feasible setpoints for the centroids $\mathcal{R}_i \subseteq \bar{\mathcal{R}}_i$, and a local control law $u_{i,k} = \kappa_i(x_{i,k}, \bar{r}_i)$. A set $\Gamma_i \subset \mathbb{R}^{n_x} \times \mathbb{R}^{d}$ is an invariant set for centroid tracking for the robot dynamics \eqref{eq:agent_dynamics} if for all $(x_{i,k}, \bar{r}_i) \in \Gamma_i$, we have $(x_{i,k}, \kappa_i(x_{i,k}, \bar{r}_i)) \in \mathcal{Z}$, $\bar{r}_i \in \mathcal{R}_i$, and $(x_{i,k+1}, \bar{r}_i) \in \Gamma_i$.
\end{definition}

Intuitively, this set represents the set of initial states and the artificial reference centroids $(x_i, \bar{r_i})$ that provides an admissible evolution of the robot dynamics defined by \eqref{eq:agent_dynamics} controlled by $u_{i,k} = \kappa_i(x_{i,k}, \bar{r}_i)$. Furthermore, once the system enters this set, it does not leave this set. The centroid tracking nonlinear constrained optimization problem, which simultaneously tracks the Voronoi centroids while minimizing bearing errors, is defined as:
\begin{subequations}
\label{eq:mpc}
\begin{align}
    \min_{u_i, \bar{x}_i, \bar{u}_i, \bar{r}_i} &J_i(x_{i,k}, r_{i,k}, g_{i,k}; \mathbf{u}_i, \bar{r}_i)\\
    &\text{s.t.} \quad l = \{1,\dots, N-1\}\\
    &x_{i,0|k} = x_{i,k}\\
    &x_{i,l+1|k} = f_i(x_{i,l|k}, u_{i,l|k})\\
    &(x_{i,l|k}, u_{i,l|k}) \in \mathcal{Z}_i\\
    &(x_{i,N|k}, \bar{r}_{i}) \in \Gamma_i\\
    &\bar{r}_i = C_i\bar{x}_i\\
    &\bar{x}_i = f_i(\bar{x}_i, \bar{u}_i),
\end{align}
\end{subequations}
where $\Gamma_i$ denotes the invariant set for centroid tracking.

Each robot then locally computes and applies the control input in a receding-horizon fashion. The decentralized control law for each robot at time $k$ is given by $ \kappa_{MPC}(x_{i,k}, r_{i,k}, g_{i,k}) = u_{i,0|k}^{*} \ \forall i \in \{1,\dots, n\}$, where $u^{*}$ is the optimal solution of \eqref{eq:mpc}. Algorithm \ref{alg:resilient_mpc} summarizes the overall control design procedure. Note that the centroid reference $r_{i,k}$ and the bearing reference $g_{i,k}$ do not affect the feasibility of the optimization problem defined in \eqref{eq:mpc} because they only appear in the cost function \eqref{eq:mpc_cost}.
\begin{algorithm}[t]
\caption{Resilient Multi-robot coverage control}\label{alg:resilient_mpc}
\begin{algorithmic}[1]
\Require Robot initial positions $\mathbf{p}_{0}$ and the MRS network topology $\mathcal{G} = (\mathcal{V}, \mathcal{E})$.
\State Compute $r_{i,0} = c_i(\mathcal{W}(\mathbf{p}_0))$, $g_{i,0} = f_B(r_{i,0})$ and $e_{i,0} = \norm{p_{i,0} - r_{i,0}} \ \forall i = \{ 1, \dots, n\}$.
\For {$k =1,2\dots,$}
    \For {$i =1,2\dots,n$}
        \State Take the measurement of $x_{i,k}$.
        \State Solve the optimization problem in \eqref{eq:mpc}.
        \State Apply the first control input $u_{i,0}^{*}$.
        \If{$\norm{p_{i,k} - r_{j,k}} \leq   e_{i,k}$ and $\exists j \in \{1,\dots,n\} \ \text{s.t.} \  \norm{p_{j,k} - r_{i,k}} \leq   e_{j,k}.$} 
            \State Update $r_{i,k} = c_i(\mathcal{W}(\mathbf{p}_k))$, $g_{i,k} = f_B(r_{i,k})$ and $e_{i,k} = \norm{p_{i,k} - r_{i,k}}$
        \EndIf 
    \EndFor
\EndFor
\end{algorithmic}
\end{algorithm}

Consider the following assumptions prevalent in the nonlinear tracking MPC literature that ensure recursive feasibility and convergence guarantees \cite{ferramosca2009mpc, limon2008mpc, limon2018nonlinear}. 
The following assumptions must be satisfied by the stage cost function, reference centroid tracking cost, and the bearing maintenance cost:
\begin{assumption}
\label{ass:cost}
\begin{enumerate}
    \item The stage cost function must satisfy $\ell_i(z,v) \geq \alpha_{\ell}(\norm{z})$ for all $(z,v) \in \mathbb{R}^{n_x+n_u}$ for some class $\mathcal{K}_\infty$ function $\alpha_{\ell}(\cdot)$.
    \item The set $\mathcal{R}_i$, which represents the set of feasible centroids, is a convex set of $\bar{\mathcal{R}_i}$.
    \item The reference centroid tracking cost $\ell_{i,r}(r_{i,k}-\bar{r}_{i})$ and the bearing maintenance cost $\ell_{i,b}(g_{i,k},\bar{r}_{i})$ denote a convex, positive semi-definite and sub-differential function such that the minimizer of the convex combination of the two satisfies:
    \begin{equation}
    \label{eq:offset_cost}
        \bar{r}_i^{\dagger}= \argmin_{\bar{r}_i \in \mathcal{R}_i} \mu \ell_{i,r}(r_{i,k} - \bar{r}_i) + (1-\mu)\ell_{i,b}(g_{i,k}, \bar{r}_i)
    \end{equation}
    is unique. Moreover, there exist $\alpha_r(\cdot)$ and $\alpha_b(\cdot)$, both class $\mathcal{K}_\infty$ functions such that
    \begin{align*}
        \ell_{i,r}(r_{i,k}-\bar{r}_{i}) -  \ell_{i,r}(r_{i,k}-\bar{r}^*_{i}) &\geq \alpha_r(\norm{\bar{r}_i - \bar{r}^*_i}),\\
        \ell_{i,b}(g_{i,k},\bar{r}_{i}) -  \ell_{i,b}(g_{i,k},\bar{r}^*_{i}) &\geq \alpha_b(\norm{\bar{r}_i - \bar{r}^*_i}).
    \end{align*}
\end{enumerate}
\end{assumption}

To ensure closed-loop stability, the terminal ingredients must follow the following assumptions:
\begin{assumption}
\label{ass:terminal}
\begin{enumerate}
    \item For the robot dynamics $x_{i,k+1}=f_i(x_{i,k}, \kappa_i(x_{i,k}, \bar{r}_i))$, $\Gamma_i$ is an invariant set for centroid tracking.
    \item Let $\kappa_i(x_{i}, \bar{r}_i)$ be the control law for all $(x_{i}, \bar{r}_i) \in \Gamma_i$, the steady-state $\bar{x}_i$ is an asymptotically stable equilibrium state for the robot dynamics $x_{i,k+1}=f_i(x_{i,k}, \kappa_i(x_{i,k}, \bar{r}_i))$. Futhermore, $\kappa_i(x_i, \bar{r}_i)$ is continuous at $(x_i, \bar{r}_i)$ for all $\bar{r}_i \in \mathcal{R}_i$.
    \item Let $\ell_{i,N}(x_{i} - \bar{x}_i, \bar{r}_i)$ be a Lyapunov function for the robot dynamics $x_{i,k+1}=f_i(x_{i,k}, \kappa_i(x_{i,k}, \bar{r}_i))$ such that for all $(x_i, \bar{r}_i) \in \Gamma_i$, there exist constants $b>0$ and $\rho >0$ such that
    \begin{align}
    \label{eq:terminal_lyap_1}
        \ell_{i,N}(x_{i} - \bar{x}_i, \bar{r}_i) &\leq b\norm{x_i - \bar{x}_i}^\rho,\\
        \ell_{i,N}(f_i(x_{i,k}, \kappa_i(x_{i,k}))& - \bar{x}_i, \bar{r}_i) -  \ell_{i,N}(x_{i,k}-\bar{x}_i, \bar{r}_i)\nonumber \\
        \label{eq:terminal_lyap_2}
        \leq -\ell_{i}(x_{i,k}-\bar{x}_i, &\kappa_i(x_{i,k}, \bar{r}_i) - \bar{u}_{i}).
    \end{align}
\end{enumerate}
\end{assumption}

These assumptions require a local controller that asymptotically stabilizes the system to any steady-state position contained in $\mathcal{R}_i$. There are several existing methods for designing such a controller in the literature, such as LTV modelling \cite{limon2018nonlinear}, linearized LQR method \cite{ chen1998quasi, kohler2019nonlinear}, etc. In order to make this exposition self-contained, we will present the design of the terminal controller and terminal invariant set for centroid tracking using an LQR control design of the linearized system around the steady-state centroid position $\bar{r}_i$, which is presented in Appendix \ref{app:terminal}.    
We now present the main convergence result in Theorem \ref{th:convergence}. It ensures the closed-loop stability of the decentralized controller that converges the robot positions to their respective Voronoi centroids. When the centroid is infeasible $r_{i,k} \notin \mathcal{R}_i$, the robots converge to the closest feasible position, which satisfies the state and the input constraints $\mathcal{Z}_i$. 
\begin{theorem}
\label{th:convergence}
Suppose Assumptions \ref{ass:invariant}, \ref{ass:dynamics}, \ref{ass:cost}, and \ref{ass:terminal} are met, and each robot computes its control inputs using Algorithm \ref{alg:resilient_mpc}. Assume the initial positions of the robots to be $p_{i,0} = C_ix_{i,0} \in \mathcal{Q} \ \forall i \in \{1,\dots, n\}$ subject to the dynamics \eqref{eq:agent_dynamics}, $r_{i,k}$ be the desired centroid position and $x_{i,0} \ \forall i \in \{1,\dots, n\}$ be the initial state satisfying the constraints in \eqref{eq:mpc}. Then, the controller obtained by solving the optimization problem \eqref{eq:mpc} is stable, recursively feasible, and converges to a steady-state position such that
\begin{enumerate}
    \item If $r_{i,k}\in \mathcal{R}_i$, then $\lim_{k \rightarrow \infty} \norm{p_{i,k} - r_{i,k}} = 0 $.
    \item If $r_{i,k} \notin \mathcal{R}_i$, then $\lim_{k \rightarrow \infty} \norm{p_{i,k} - \bar{r}_{i}^{\dagger}} = 0 $, where
    \begin{equation*}
         \bar{r}_i^{\dagger}= \text{arg} \min_{\bar{r}_i \in \mathcal{R}_i} \mu \ell_{i,r}(r_{i,k} - \bar{r}_i) + (1-\mu)\ell_{i,b}(g_{i,k}, \bar{r}_i). 
    \end{equation*}
\end{enumerate}   
\end{theorem}
\begin{proof}
The proof is presented in Appendix \ref{app:convergence}.    
\end{proof}
\subsection{Rigidity Recovery under Agent Loss}
We will now show how maintaining bearing rigidity allows us to design a systematic procedure to recover an MRS network in the case of robot loss. Moreover, we want to recover the network topology to remain minimally rigid even after reconfiguration. This will ensure that even after the reconfiguration, the robots maintain a unique topology with a minimum number of links, thereby ensuring resiliency with minimum additional connections. In rigidity theory literature, such problems are called \textit{minimal-rigidity preserving closing ranks} problem. This can be formally defined as follows. 

Given a minimally rigid graph $\mathcal{G} = (\mathcal{V}, \mathcal{E})$ and let  $\mathcal{G}_f = (\mathcal{V} \backslash \{v_r\}, \mathcal{E} \backslash \mathcal{E}_r)$ be the graph obtained when a robot representing $v_r \in \mathcal{V}$ and all edges incident to $v_r$ denoted as $\mathcal{E}_r$ are removed from the graph $\mathcal{G}$. The goal is to find a minimum set of new edges $\mathcal{E}_n$ which must be added to $\mathcal{G}_f$ to yield it minimally rigid. We now present an important result that establishes the existence and lower bounds on the number of edges to be added to yield a minimal rigidity.
\begin{lemma}[Local Repair, Th. 3\cite{fidan2010closing}]
\label{lem:closing_rank}
Consider a vertex $v$ of degree $\alpha$ being removed from a $d$-dimensional minimally rigid graph, where $d=2$ or $d=3$ and $\alpha\geq d$. To restore minimal rigidity, $\alpha - d$ edges (or none if $\alpha =d$) must be added. Furthermore, these new edges should only connect the neighbors of $v$, without using any other vertex as an endpoint.
\end{lemma}
\begin{algorithm}[t]
\caption{Rigidity Recovery using Edge contraction}\label{alg:recovery}
\begin{algorithmic}[1]
\Require MRS network topology $\mathcal{G} = (\mathcal{V}, \mathcal{E})$.
\For {$i =1,2\dots,n$}
    \For {$j =1,2\dots, \mathcal{N}_i$}
        \If{($\alpha_j$ == $d$) degree of the vertex $v_j$}
        \Return
        \Else
            \State Check whether $e_{i,p} = (v_i,v_p) \ \forall p \in \{1, \dots, \mathcal{N}_j\}$ is contractible or not using Lemma \ref{lem:non_contract}.
            \State $\mathcal{E}^r_{ij} \leftarrow \big\{e_{i,q}, q\big\}  \ \forall q \in \{1, \dots, \mathcal{N}_j\}$ such that $e_{i,q}$ is contractible. 
        \EndIf
    \EndFor
\EndFor
\end{algorithmic}
\end{algorithm}

\begin{figure}[h]
\centering
\includegraphics[width=0.5\textwidth, trim={0.5cm 0.5cm 0.5cm 0.75cm}]{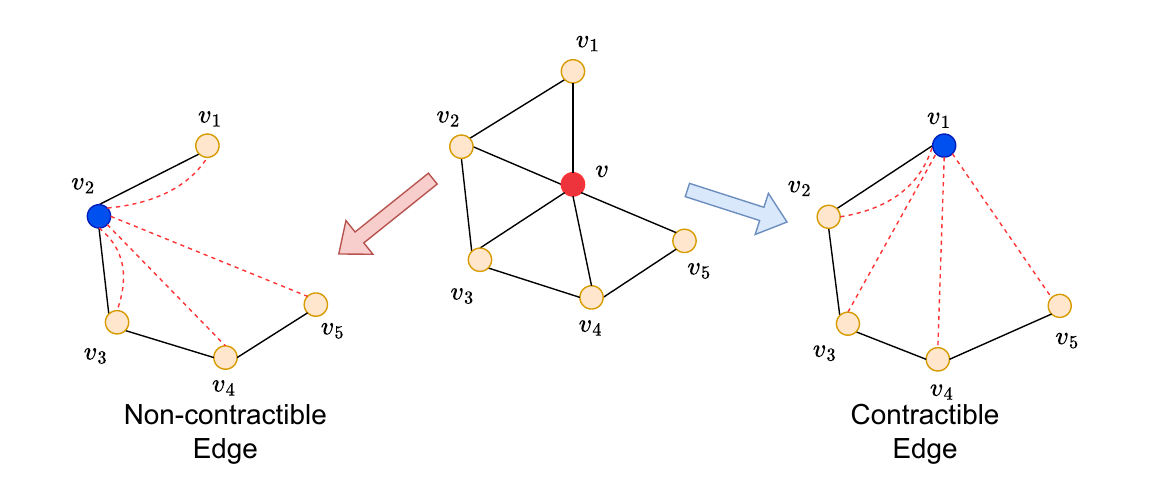}
\caption{Edge contraction operation for Rigidity recovery.}
\label{fig:closing_rank}
\end{figure}

While the above lemma illustrates how to regain graph rigidity when a vertex is lost, it doesn't provide a constructive approach for finding the minimum edge set required to regain rigidity. A systematic procedure to find the minimum edge set can be obtained by using \textit{edge contraction approach} \cite{fidan2010closing}. The idea is to merge two adjacent vertices $v_1,v_2 \in \mathcal{V}$ of a graph $\mathcal{G} = (\mathcal{V}, \mathcal{E})$ into a single vertex by ``contracting" the edge between them. This new vertex should be adjacent to all the neighbors of $v_1$ and $v_2$ without redundant edges (if any would exist). If the resulting graph after the edge contraction operation on an edge $e$ is minimally rigid, then the edge $e$ is called a contraction edge. Figure \ref{fig:closing_rank} shows the edge contraction operation on a minimally rigid graph. We can see in the right side of the figure that when the vertex $v$ (in red) is removed in a minimally rigid graph, we can find new edges $(v_1,v_3), (v_1, v_4), (v_1, v_5)$ using edge contraction operation on $(v, v_1)$. The resulting graph is not minimally rigid when the edge contraction is performed on $(v, v_2)$. Hence, we get a minimally-rigid graph. The sufficient condition to check whether a given edge is contractible \cite{fidan2010closing} is given by the following lemma:
\begin{lemma}[Edge Contraction\cite{fidan2010closing}]
\label{lem:non_contract}
Consider a minimally rigid graph $\mathcal{G} = (\mathcal{V}, \mathcal{E})$ in $d=2$. Let $v,w \in \mathcal{V}$ and $(v,w) \in \mathcal{E}$. The edge $(v,w)$ is non-contractible in $\mathcal{G}$ if multiple vertices (more than one vertex) are adjacent to both $v$ and $w$.
\end{lemma}

Using the results from Lemmas \ref{lem:closing_rank} and \ref{lem:non_contract}, we design a proactive rigidity recovery algorithm as shown in Alg. \ref{alg:recovery}. Using the results from Lemma \ref{lem:closing_rank}, we are only required to search locally for the contractible edges among the $2$-hop neighbors of the robot. The algorithm checks all possible contractible edges for the neighbor of each robot in the MRS network. The idea is to compute recovery edge sets, $\mathcal{E}^r_{ij}: \big\{ \{(i,j), q\} | \ i,j \in \mathcal{V}, q \in \{i,j\} \big\} \quad \forall i \in \{1,\dots,n\},  \ \forall j \in \mathcal{N}_i$, comprising new recovery edges $\big\{(i,j), \ i,j \in \mathcal{V}\big\}$ and the corresponding contraction vertex $q \in \{i,j\}$ for each robot $i$ by anticipating a failure in each of the neighbors $j$. This ensures that each robot can reconfigure itself by making new connections with the help of new recovery edges in the event of robot loss \footnote{Note that these results are applicable only for $d=2$. The results from the planar case do not scale well for higher dimensions \cite{fidan2010closing}.}. Moreover, the set of recovery edges also guarantees a minimally rigid graph after reconfiguration. 
\begin{figure*}[t]
\centering
        \centering
        \includegraphics[width=0.8\textwidth, trim={2.8cm 1.5cm 1.5cm 1.5cm},clip]{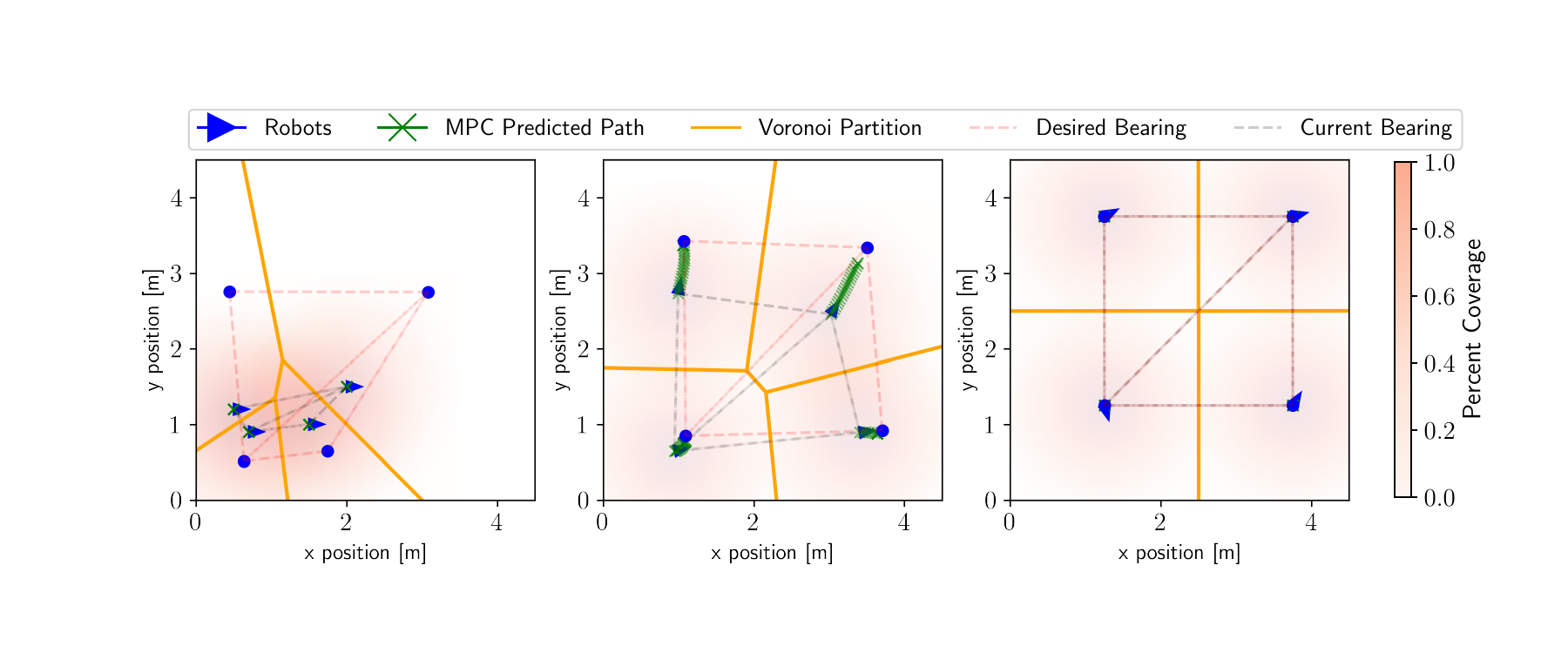}
\caption{Robot positions and Voronoi partitions along with bearing formation at time $k=0$, $50$, $140$.}
\label{fig:coverage}
\end{figure*}
\begin{figure}[ht]
\centering
\includegraphics[width=0.5\textwidth,trim={0.5cm 0.4cm 0.3cm 0.2cm},clip]{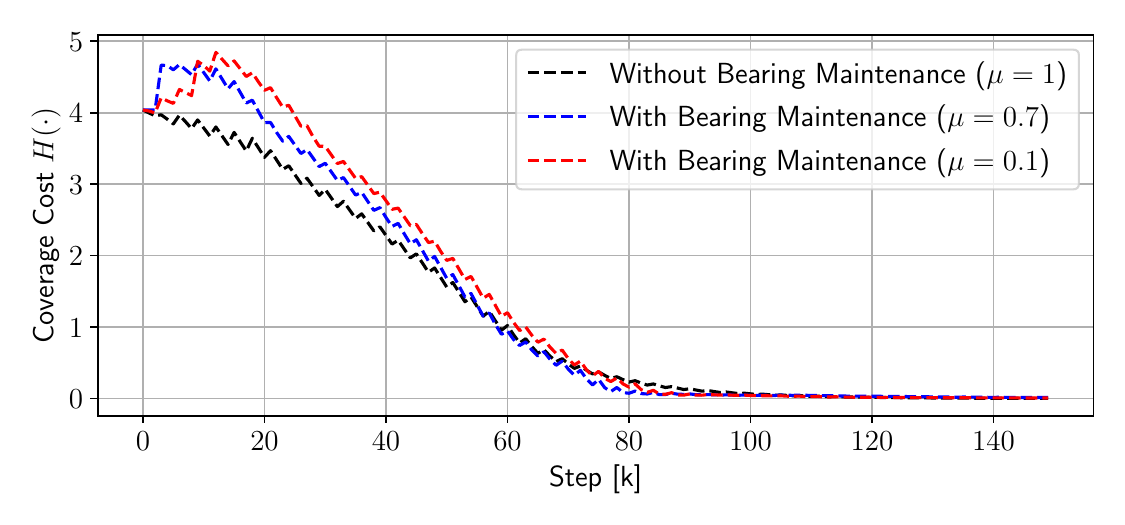}
\caption{Coverage cost evolution with time, for three different values of $\mu$: (a) without bearing maintenance $(\mu = 1)$ and (b) with bearing maintenance $(\mu = 0.7)$ and $(\mu = 0.1)$.}
\label{fig:cost}
\end{figure} 
\section{Numerical Simulations}
\label{sec:sim}
In this section, we show the effectiveness of our proposed resilient coverage control and rigidity recovery algorithms using numerical simulations. 
\begin{figure*}[h]
\centering
        \centering
        \includegraphics[width=0.9\textwidth, trim={2.8cm 2.3cm 1.5cm 2.6cm},clip]{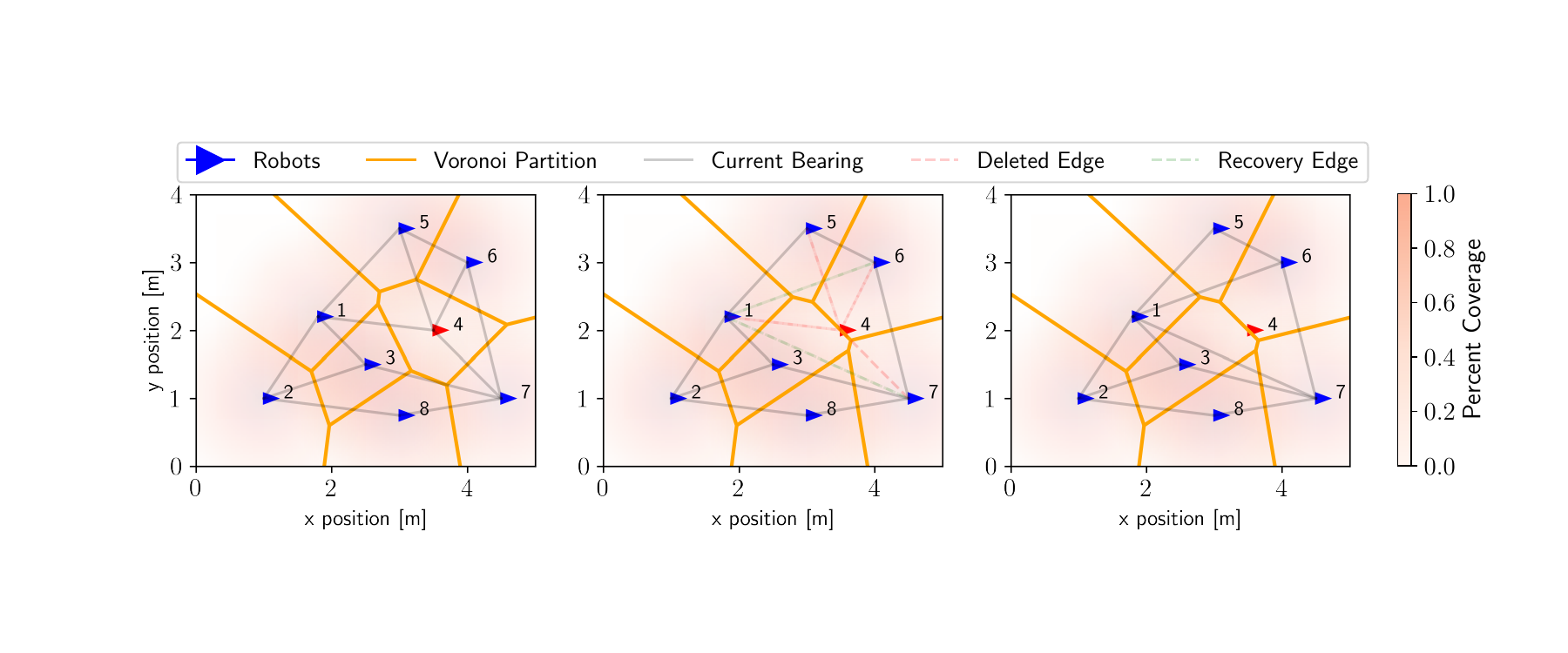}
\caption{Rigidity recovery in the event of loss of robot $4$. The neighbors of robot $4$, i.e., robots 1,5, 6, and 7, utilize the proposed rigidity recovery algorithm to identify $(1,6)$ and $(1,7)$ as the new set of edges for rigidity recovery.}
\label{fig:rigidity_recovery}
\end{figure*}

\textit{Results - Bearing Maintenance in Coverage Control}: The coverage environment is defined as $\mathcal{Q} = [0,5] \times [0,5]$. The density function is chosen as $\phi(q) = 1$. We consider $4$ robots, each modeled as the unicycle dynamics. The dynamics can be represented as,
\begin{align}
    p^x_{i,k+1} &= p^x_{i,k} + T_s \cos (\theta_{i,k}) v_{i,k}\\
    p^y_{i,k+1} &= p^y_{i,k} + T_s \sin (\theta_{i,k}) v_{i,k}\\
    \theta_{i,k+1} &= \theta_{i,k} + T_s \omega_{i,k},
\end{align}
where $p_{i,k}^x$ and $p_{i_k}^y$ denote the positions and $\theta_{i,k}$ denotes the yaw of the $i^{\text{th}}$ robot at time $k$, respectively. The control inputs are the linear and angular velocities denoted by $v_{i,k}$ and $\omega_{i,k}$ for each robot, respectively. The sample time is $T_s = 0.1s$. The matrix $C_i$ is given by $
C_i = \begin{bsmallmatrix}
    1 & 0 & 0 & 0\\
    0 & 1 & 0 & 0
\end{bsmallmatrix}$. 

The position of each robot is constrained to be within the coverage region $\mathcal{Q}$ and the yaw $\theta_{i,k} \in [-\pi, \pi]$. The linear and angular velocities for each robot are limited to $\norm{v_{i}}\leq 1$ $m/s$ and $\norm{\omega_{i}}\leq \pi/2$ $rad/s$, respectively. The initial positions of all the robots are $p_{1,0} = [0.7,0.9]^\top$, $p_{2,0} = [0.5,1.2]^\top$, $p_{3,0} = [2.0,1.5]^\top$, and $p_{4,0} = [1.5, 1.0]^\top$. The stage cost is given by $\ell_i(x_i-\bar{x}_i,u_i - \bar{u}_i) = \norm{x_i - \bar{x}_i}^2_{Q_i} + \norm{u_i - \bar{u}_i}^2_{R_i}$, where $Q_i = \text{diag}(1,1,1)$ and $R_i= \text{diag}(0.05, 0.005)$ are chosen as diagonal matrices. The reference tracking cost and bearing maintenance costs are given by $\ell_{i,r}(r_i-\bar{r}_i) = \norm{r_i - \bar{r}_i}^2_{Q^r_i}$ and $\ell_{i,b}(g_i, \bar{r}_i) = \sum_j^{\mathcal{N}_i} \norm{g_{ij} - f_B(\bar{r}_i)}^2_{Q^b_{ij}}$, respectively, where $Q^r_i = I$ and $Q^b_{ij} = 100\cdot I$. 

Figure \ref{fig:coverage} shows the evolution of the robots with the initial configuration at $k = 0$, the intermediate configuration at $k=50$ and the final configuration $k=140$. We can clearly observe that our proposed algorithm converges to the $4$ rectangles of equal area as $k$ becomes $140$, validating our theoretical results. In Fig. \ref{fig:cost}, we plot the coverage cost function \eqref{eq:coverage_prob} with time until $k =140$ for different values of $\mu$. By modifying the value of $\mu$, we can tune the importance of the bearing cost on the overall optimization problem. In Fig. \ref{fig:cost}, we compare the coverage cost for three different values of $\mu=1$, $\mu = 0.7$ and $\mu = 0.1$. We observe that as the weights on the bearing cost increase, i.e., $1 - \mu = 0.3$ and $1 - \mu = 0.9$, the coverage cost has a nonmonotonic behavior at the beginning. This is because our proposed controller first aligns the robots to enforce a constant bearing formation and then translates toward the desired centroids\footnote{Note that when $\mu=1$, our algorithm is equivalent to \cite{carron2020model}}. Thus, our numerical simulations corroborate and validate the main convergence result.

\textit{Results - Rigidity Recovery Under Agent Loss}: To illustrate the effectiveness of the proposed rigidity recovery algorithm, we consider a network of $8$ robots assigned to accomplish a coverage task as shown in Fig. \ref{fig:rigidity_recovery}. We consider that robot $4$ becomes faulty during the mission and leaves the network. Thus, the rest of the robots in the network must reorganize themselves to ensure connectivity (rigidity) and, thus, the mission's objective. Using Algorithm \ref{alg:recovery}, each of the neighbors of robot $4$, i.e., robots 1, 5, 6 and 7, proactively computes the recovery set $\mathcal{E}^r_{i4}, \ i\in\{1,5,6,7\}$ with $j=4$ in the following way:
\begin{align*}
    \mathcal{E}^r_{14} &= \big\{\{(1,6), 1\}, \{(1,7), 1\}\big\}, \quad \mathcal{E}^r_{54} = \big\{\{(5,7), 7\}\big\}, \\
    \mathcal{E}^r_{64} &= \big\{\{(1,6), 1\} \big\}, \quad \mathcal{E}^r_{74} = \big\{\{(1,7), 1\}, \{(5,7), 7\}\big\}.
\end{align*}
As a result, upon the failure of robot $4$, the robot network reconfigures itself by introducing $(1,6)$ and $(1,7)$ using robot $1$ as the contraction vertex in the underlying MRS network. Note that the resulting network after reconfiguration results in $2*7-3 = 11$ edges, thus forming a Laman graph, i.e., minimally bearing rigid graph. This illustrates that by ensuring bearing maintenance during coverage tasks, the network can be recovered in case of a robot loss.

\section{Conclusion}
\label{sec:conc}
In this paper, we proposed a resilient multi-robot coverage control algorithm that enhances the structural resilience of the MRS network during the coverage task. By posing the coverage problem as scaling and translating a virtual rigid structure formed by robots and their communication links, we showed that the resiliency of the underlying network could be enhanced in the presence of faults/cyberattacks or robot loss. We formulated the simultaneous centroid tracking and rigidity maintenance problem as an output tracking nonlinear MPC problem and proved that our proposed algorithm converges to the centroidal Voronoi configuration. Furthermore, as a consequence of bearing rigidity, we designed a rigidity recovery algorithm to reconfigure the MRS network in the event of a robot failure, which computes the set of new recovery edges for the neighbors of the faulty robot, enhancing the self-healing capabilities of the underlying MRS network. Finally, we conducted numerical simulations to corroborate our theoretical results. In the future, we would like to extend our approach to reconfigure the network despite multiple simultaneous robot failures when deployed in a dynamic environment.  



\appendices
\section{Convergence Results and Proofs}
\label{app:convergence}
\subsection{Proof of Theorem \ref{th:convergence}}
The proof is divided into two parts. We first prove the recursive feasibility of the optimization problem throughout the evolution of the robot dynamics. We then prove the closed-loop stability and optimality of the steady-state position $\bar{r}_i^\dagger$ and its corresponding steady-state state-input pair $(\bar{x}_i^\dagger, \bar{u}_i^\dagger)$. For brevity, we define $\ell_{i,c}^\mu(r_{i,k},g_{i,k},\bar{r}_i) := \mu \ell_{i,r}(r_{i,k} - \bar{r}_i) + (1-\mu)\ell_{i,b}(g_{i,k}, \bar{r}_i)$ for the rest of the proof.

We first introduce the following lemma, which will play an important role in proving the stability of the proposed decentralized centroid tracking controller. 
\begin{lemma}
\label{lem:stab_eq}
Suppose Assumptions \ref{ass:invariant}, \ref{ass:dynamics}, \ref{ass:cost}, and \ref{ass:terminal} are met, and each robot dynamics is governed by \eqref{eq:agent_dynamics} with constraints $(x_{i,k},u_{i,k})\in \mathcal{Z}_i$. Consider for a given position setpoint, determined by Voronoi centroid $r_{i,k}$, the optimal steady-state solution of the MPC cost function $J_i^*(x_{i,k}, r_{i,k}, g_{i,k}; \mathbf{u}_i, \bar{r}_i)$ at time $k$ is given by $(\bar{x}_i^*, \bar{u}_i^*, \bar{r}_i^*)$. Then, when $x_{i,k} = \bar{x}_i^*$, the optimal cost function
$J_i^*(x_{i,k}, r_{i,k}, g_{i,k}; \mathbf{u}_i, \bar{r}^{\dagger}_i) = \ell_{i,c}^\mu(r_{i,k},g_{i,k},\bar{r}^\dagger_i)$
where $\bar{r}_i^{\dagger}$ is same as \eqref{eq:offset_cost}.
\end{lemma}
\begin{proof}
Let the optimal steady-state solution of the MPC optimization cost $J_i^*(x_{i,k}, r_{i,k}, g_{i,k}; \mathbf{u}_i, \bar{r}_i)$ at time $k$ be denoted as $(\bar{x}_i^*, \bar{u}_i^*, \bar{r}_i^*)$. By the definition of the MPC cost function \eqref{eq:mpc_cost}, its optimal value, when $x_{i,k} = \bar{x}_i^*$ is, 
\begin{equation}
    J_i^*(\bar{x}_{i}^*, r_{i,k}, g_{i,k}) =\ell_{i,c}^\mu(r_{i,k},g_{i,k},\bar{r}_i^*).
\end{equation}
We will use proof by contradiction to prove this result. Let us assume that for a given $\mu \in (0, 1 ]$, $\ell_{i,c}^\mu(r_{i,k},g_{i,k},\bar{r}_i^*) > \ell_{i,c}^\mu(r_{i,k},g_{i,k},\bar{r}_i^\dagger)$, this implies that $\bar{r}_i^\dagger$ is the unique minimizer of the combined cost function $\ell_{i,c}^\mu(\cdot)$, and $\bar{r}_i^\dagger\neq \bar{r}_i^*$.
Let us define $\hat{r}_i$ as
\begin{equation}
\label{eq:convex_set_ref}
    \hat{r}_i = \beta \bar{r}_i^* + (1-\beta)\bar{r}_i^\dagger, \quad \beta \in [0,1].
\end{equation}
Note that by the definition of $\mathcal{R}_i$, the state-input pair satisfies $(\bar{x}_i^*, \bar{u}_i^*) \in \hat{\mathcal{Z}_i}$ and hence lies in the interior of the constraint set $\mathcal{Z}_i$.
Therefore, there exists a $\hat{\beta} \in [0,1)$, as a consequence of the restricted set $\bar{\mathcal{Z}_i}$ defined earlier, such that for a given $\beta \in [\hat{\beta}, 1]$, it is true that $(\bar{x}^*_{i}, \hat{r}_i) \in \Gamma_i$. That is, the state and the reference centroid lie in the invariant set for centroid tracking, and the sequence of inputs $\hat{\mathbf{u}}_i$ (generated by $\kappa_i(\bar{x}^*_{i}, \hat{r}_i)$) and $\hat{r}_i$ are a feasible solution of \eqref{eq:mpc}. 

Using the optimality of the solution $J_i^*(\cdot)$ and Assumption \ref{ass:terminal} providing bounds on the summation of the stage costs, the following holds:
\begin{align}
\label{eq:stage_cost_bound}
    \ell_{i,c}^\mu(r_{i,k},&g_{i,k},\bar{r}_i^*) = J_i^*(\bar{x}_{i}^*, r_{i,k}, g_{i,k}) \nonumber
    \\&\leq J_i(\bar{x}_{i}^*, r_{i,k}, g_{i,k}; \hat{\mathbf{u}}_i, \hat{r}_i)\nonumber
    \\&=\sum_{l=0}^{N-1} \ell_{i,l}(x_{i,l|k} - \bar{x}_{i}, \kappa_i(\bar{x}_i^*, \hat{r}_i) -  \bar{u}_i) \nonumber
    \\& + \ell_{i,N}(x_{i,N|k} - \bar{x}_i, \hat{r}_i)+ \ell_{i,c}^\mu(r_{i,k},g_{i,k},\hat{r}_i)
\end{align}
\begin{align}
\label{eq:stage_cost_bound_2}
    \ell_{i,c}^\mu(r_{i,w},g_{i,w},\bar{r}_i^*) &\stackrel{\text{(a)}} \leq \ell_{i,N}(\bar{x}^*_{i} - \hat{x}_i, \hat{r}_i) + \ell_{i,c}^\mu(r_{i,w},g_{i,w},\hat{r}_i)\nonumber
    \\\stackrel{\text{(b)}}\leq& b|\bar{x}^*_{i} - \hat{x}_i|^\rho + \ell_{i,c}^\mu(r_{i,w},g_{i,w},\hat{r}_i)\nonumber
    \\\stackrel{\text{(c)}}\leq& b \frac{\norm{\bar{r}_i^* - \hat{r}_i}^\rho}{\bigl(\sigma_{min}(C_i)\bigr)^\rho} + \ell_{i,c}^\mu(r_{i,w},g_{i,w},\hat{r}_i)\nonumber
    \\\stackrel{\text{(d)}}\leq& b \frac{(1-\beta)^\rho\norm{\bar{r}_i^* - \bar{r}_i^\dagger}^\rho}{\bigl(\sigma_{min}(C_i)\bigr)^\rho} + \ell_{i,c}^\mu(r_{i,w},g_{i,w},\hat{r}_i)
\end{align}
Step $\text{(a)}$ is a result of the summation of \eqref{eq:terminal_lyap_2} on both sides, resulting in the cancellation of alternate terms in the LHS yielding the inequality. For step $\text{(b)}$, we use the property of the terminal cost function \eqref{eq:terminal_lyap_1}. In step $\text{(c)}$, we use the property of the singular value of matrices, $\norm{\bar{x}_i^* - \hat{x}_i} \leq \frac{1}{\sigma_{min}(C_i)}\norm{\bar{r}_i^* - \hat{r}_i}$, where $\sigma_{min}(\cdot)$ is the minimum singular value of a matrix. Step $\text{(d)}$ is obtained by rearranging terms in \eqref{eq:convex_set_ref} and taking norm on both sides, i.e,  $\norm{\bar{r}_i^* - \hat{r}_i} = (1-\beta)\norm{\bar{r}_i^* - \bar{r}_i^\dagger}$.

By using the convexity of $\ell_{i,r}(\cdot)$ and $\ell_{i,b}(\cdot)$ with \eqref{eq:convex_set_ref}, we have
\begin{align}
     \ell_{i,c}^\mu(r_{i,k},g_{i,k},\hat{r}_i) \leq \ &\beta \ell_{i,c}^\mu(r_{i,k},g_{i,k},\bar{r}_i^*) \nonumber
     \\&+ (1-\beta)\ell_{i,c}^\mu(r_{i,k},g_{i,k},\bar{r}_i^\dagger).
\end{align}
Thus, we can derive the following inequality
\begin{align}
      \ell_{i,c}^\mu(r_{i,k},&g_{i,k},\bar{r}_i^*) \leq b \frac{(1-\beta)^\rho|\bar{r}_i^* - \bar{r}_i^\dagger|^\rho}{\bigl(\sigma_{min}(C_i)\bigr)^\rho} \nonumber
      \\&\beta \ell_{i,c}^\mu(r_{i,k},g_{i,k},\bar{r}_i^*) + (1-\beta)\ell_{i,c}^\mu(r_{i,k},g_{i,k},\bar{r}_i^\dagger)
\end{align}
By rearranging terms, we get
\begin{align}
      \ell_{i,c}^\mu(r_{i,k},g_{i,k},&\bar{r}_i^*) - \ell_{i,c}^\mu(r_{i,k},g_{i,k},\bar{r}_i^\dagger) \nonumber
      \\ &\leq \frac{b(1-\beta)^{\rho-1}\norm{\bar{r}_i^* - \bar{r}_i^\dagger}^\rho}{\bigl(\sigma_{min}(C_i)\bigr)^\rho}
\end{align}~
Since $\rho>1$, taking limit on both sides as $\beta \rightarrow 1^-$ from the right, we get
\begin{align}
\label{eq:lemma1_cont}
      \ell_{i,c}^\mu(r_{i,k},g_{i,k},\bar{r}_i^*) - \ell_{i,c}^\mu(&r_{i,k},g_{i,k},\bar{r}_i^\dagger) \nonumber
      \\ &\leq \lim_{\beta\rightarrow1^-} \frac{b(1-\beta)^{\rho-1}\norm{\bar{r}_i^* - \bar{r}_i^\dagger}^\rho}{\bigl(\sigma_{min}(C_i)\bigr)^\rho}\nonumber
       \\& = 0
\end{align}~
Since we initially assumed that the combined cost follows $\ell_{i,c}^\mu(r_{i,k},g_{i,k},\bar{r}_i^*) - \ell_{i,c}^\mu(r_{i,k},g_{i,k},\bar{r}_i^\dagger)>0$, the inequality \eqref{eq:lemma1_cont} leads to contradiction which proves the result.
\end{proof}~
We will now prove the theorem by first proving the recursive feasibility of the proposed approach.

\textit{Recursive Feasibility}: Let us consider the state of robot $i$ at time $k$ to be $x_{i,k}$, and the optimal solution of the nonlinear centroid tracking control problem \eqref{eq:mpc} with $x_{i,k}$ and $r_{i,k}$ as inputs, is given by $(\mathbf{u}_i^*, \bar{r}_i^*)$. The resulting optimal sequence of the state is given by
\begin{equation}
    \mathbf{x}_i^* = [(x_{i,0|k}^*)^\top, (x_{i,1|k}^*)^\top, \dots, (x_{i,N|k}^*)^\top]^\top,
\end{equation}
where $(x_{i,N|k}^*, \bar{r}_i^*) \in \Gamma_i$.

To prove the recursive feasibility, we use identical steps from the standard MPC literature \cite{ferramosca2009mpc,limon2018nonlinear,limon2008mpc}, we define the successor state, $x_{i,k+1} = f_i(x_{i,k}, \kappa_{MPC}(x_{i,k}, r_{i,k}))$. We also define the following sequences:
\begin{align}
\label{eq:inp_seq}
    \mathbf{u}_i^+ &= [(u_{i,1|k}^*)^\top, (u_{i,2|k}^*)^\top, \dots, (u_{i,N-1|k}^*)^\top, \kappa_i(x_{i,N|k}^*, \bar{r}_i^*)]^\top, \nonumber
    \\ r_i^+ &=  \bar{r}_i^*
\end{align}
Note that the first $N-1$ entries of the sequence $\mathbf{u}_i^+$ are the optimal control sequence from the optimization step at time $k$. It is augmented by an additional control input using the terminal control law $\kappa_i(\cdot)$. The predicted sequence of states for $x_{i,k+1}$ by choosing the control sequence $\mathbf{u}_i^+$ and $ r_i^+$ as reference centroid is given by:
\begin{align}
\label{eq:state_seq}
    \mathbf{x}_i^+ = [(x_{i,1|k}^*)^\top, \dots, (x_{i,N|k}^*)^\top, (x_{i,N|k+1})^\top]^\top
\end{align}
where $x_{i,N|k+1}$ denotes the terminal state at time $k+1$, given by $x_{i,N|k+1} = f_i(x_{i,N-1|k+1}, \kappa_i(x_{i,N-1|k+1}, \bar{r}_i^*))$. Using the sequence of states in \eqref{eq:state_seq}, we observe that when applying the control sequence $\mathbf{u}_i^+$ at time $k+1$, the following holds: $x_{i,N-1|k+1} = x_{i,N|k}^*$.
As a result, since we have $(x_{i,N|k}^*, \bar{r}_i^*) \in \Gamma_i$, then $(x_{i,N|k+1}^*, \bar{r}_i^*)\in \Gamma_i$.

Therefore, given $x_{i,1|k}^* = x_{i,k+1}$, the control sequence and reference centroid position defined by $(\mathbf{u}_i^+,r_i^+)$ are a feasible solution of \eqref{eq:mpc}. Hence, the proposed control law is well-defined and yields constraint satisfaction consistently throughout the evolution of the robot's dynamics.

\textit{Asymptotic Stability}: We will prove this by showing that the steady-state state-input pair $(\bar{x}_i^\dagger, \bar{u}_i^\dagger)$ corresponding to the minimizer of the combined reference tracking and bearing cost (i.e., $\bar{r}_i^\dagger$) is a stable equilibrium point and it is attractive for the robot dynamics \eqref{eq:agent_dynamics}. To show the stability of $(\bar{x}_i^\dagger, \bar{u}_i^\dagger)$, we demonstrate that the following function defined as
\begin{equation*}
      W_i(x_{i,k}, r_{i,k}, g_{i,k}) = J_i^*(x_{i,k}, r_{i,k}, g_{i,k})-\ell_{i,c}^\mu(r_{i,k},g_{i,k},\bar{r}_i^\dagger) 
\end{equation*}
is a Lyapunov function for the closed-loop robot dynamics in the neighbourhood of the equilibrium point.
The terminal control law $u_i = \kappa_i(x_{i,k}, \bar{r}_i^\dagger)$ is admissible for all $x_{i,k}$ satisfying $\norm{x_{i,k} - \bar{x}_i^\dagger}\leq \epsilon$, where $\epsilon>0$ can be chosen as a sufficiently small value to ensure that $x_{i,k}$ is in the neighborhood of $\bar{x}_i^\dagger$.
We now prove that $W_i(x_{i,k}, r_{i,k}, g_{i,k})$ is a locally positive-definite function by proving the existence of a pair of suitable class $\mathcal{K}_\infty$ functions, $\alpha_W(\cdot)$ and $\beta_W(\cdot)$, such that 
\begin{equation*}
    \alpha_W(\norm{x_{i,k} - \bar{x}_i^\dagger}) \leq W_i(x_{i,k}, r_{i,k}, g_{i,k}) \leq \beta_W(\norm{x_{i,k} - \bar{x}_i^\dagger}).
\end{equation*}
Using Assumption \ref{ass:cost}, we can infer that
\begin{align}
    W_i(x_{i,k}, r_{i,k}, g_{i,k})&\geq  \ \ell_i({x_{i,k} - \bar{x}_i^*,u_{i,k} - \bar{u}_i^*}) \nonumber
    \\&+ \ell_{i,c}^\mu(r_{i,k},g_{i,k}, \bar{r}_i^*) - \ell_{i,c}^\mu(r_{i,k},g_{i,k},\bar{r}_i^\dagger) \nonumber
    \\&\geq \alpha_\ell(\norm{x_{i,k} - \bar{x}_i^*}) + \mu \alpha_r(\norm{\bar{r}_i^* - \bar{r}_i^\dagger}) \nonumber
    \\&\quad + (1-\mu)\alpha_b(\norm{\bar{r}_i^* - \bar{r}_i^\dagger})\nonumber 
    \\&\geq \alpha_\ell(\norm{x_{i,k} - \bar{x}_i^*}) + \mu \hat{\alpha}_r(\norm{\bar{x}_i^* - \bar{x}_i^\dagger})\nonumber
    \\&\quad + (1-\mu)\hat{\alpha}_b(\norm{\bar{x}_i^* - \bar{x}_i^\dagger}),\nonumber
\end{align}
where $\hat{\alpha}_r(\norm{\bar{x}_i^* - \bar{x}_i^\dagger}) = \alpha_r\bigl(\sigma_{max}(C_i)\norm{\bar{x}_i^* - \bar{x}_i^\dagger}\bigr)$ and $\hat{\alpha}_b(\norm{\bar{x}_i^* - \bar{x}_i^\dagger}) = \alpha_b\bigl(\sigma_{max}(C_i)\norm{\bar{x}_i^* - \bar{x}_i^\dagger}\bigr)$ are class $\mathcal{K}_\infty$ functions with $\sigma_{max}(C_i)$ as the maximum singular value of the matrix $C_i$. Then, there exists a class $\mathcal{K}_\infty$ function $\alpha_W(\cdot)$ such that 
\begin{align}
    W_i(x_{i,k}, r_{i,w}, g_{i,w}) &\geq \alpha_W(\norm{x_{i,k} - \bar{x}_i^*} + \norm{\bar{x}_i^* - \bar{x}_i^\dagger})\nonumber\\
    &\geq \alpha_W(\norm{x_{i,k} - \bar{x}_i^\dagger}).
\end{align}
Consider the input sequence generated by the local control law using $\bar{r}_i^\dagger$ to be the centroid reference position $\mathbf{u}^\kappa_{i} = \big[\bigl(\kappa_i(x_{i,0|k},\bar{r}_i^\dagger)\bigr)^\top, \dots, \bigl(\kappa_i(x_{i,N|k},\bar{r}_i^\dagger)\bigr)^\top\big]^\top$. This sequence will be feasible and satisfies the following inequality,
\begin{align}
    J_i^*(x_{i,k}, r_{i,k}, g_{i,k}) &\leq J_i(x_{i,k}, r_{i,k}, g_{i,k}; \mathbf{u}^\kappa_i, \bar{r}_i^\dagger)\nonumber\\
    \leq \ell_{i,N}&(x_{i,k} - \bar{x}_i^\dagger, \bar{r}_i^\dagger) + \ell_{i,c}^\mu(r_{i,k},g_{i,k},\bar{r}_i^\dagger).
\end{align}
The above inequality results from the summation of \eqref{eq:terminal_lyap_2} on both sides, following step $\text{(a)}$ in \eqref{eq:stage_cost_bound}.

Using the above inequality and the conditions from Assumption \ref{ass:invariant}, we get
\begin{align}
    W_i(x_{i,k}, r_{i,k}, g_{i,k}) & \leq\ell_{i,N}(x_{i,k} - \bar{x}_i^\dagger, \bar{r}_i^\dagger) \leq b\norm{x_{i,k} - \bar{x}_i^\dagger}\nonumber\\
    &\leq \beta_W(\norm{x_{i,k} - \bar{x}_i^\dagger}).
\end{align}

Next, we have to prove that $W_i(x_{i,k}, r_{i,k}, g_{i,k})$ is a strictly decreasing function if $x_{i,k}\neq \bar{x}_{i}^*$. We define $J_i(x_{i,k+1}, r_{i,k}, g_{i,k}; \mathbf{u}_i^+, \bar{r}_i^*)$ as the cost function evaluated at the successor state $x_{i,k+1} = f_i(x_{i,k}, \kappa_{MPC}(x_{i,k}, r_{i,k}))$ with $\mathbf{u}_i^+$ as the input sequence and $\bar{r}_i^*$ as the position of the reference centroid defined in \eqref{eq:inp_seq}. Using Assumption \ref{ass:terminal} and standard procedures in deriving closed-loop stability of standard MPC \cite{limon2008mpc, limon2018nonlinear, ferramosca2009mpc}, we obtain:
\begin{align}
    \Delta&W_i(x_{i,k}, r_{i,k}, g_{i,k})\nonumber
    \\&= W_i(x_{i,k+1}, r_{i,k}, g_{i,k}) - W_i(x_{i,k}, r_{i,k}, g_{i,k})\nonumber
    \\&= J_i^*(x_{i,k+1}, r_{i,k}, g_{i,k}) - J_i^*(x_{i,k}, r_{i,k}, g_{i,k})\nonumber
    \\&\leq J_i(x_{i,k+1}, r_{i,k}, g_{i,k}; \mathbf{u}_i^+, \bar{r}_i^*) - J_i^*(x_{i,k}, r_{i,k}, g_{i,k})\nonumber
    \\&=-\ell_{i}(x_{i,k} - \bar{x}_i^*, u_{i,k} - \bar{u}_i^*)-\ell_{i,N}(x_{i,k} - \bar{x}_i^*, \bar{r}_i^*)\nonumber
    \\&-\ell_{i,c}^\mu(r_{i,k},g_{i,k},\bar{r}_i^*) +\ell_{i}(x_{i,N|k} - \bar{x}_i^*, \kappa_i(x_{i,k}, \bar{r}_i^*) - \bar{u}_i^*)\nonumber
    \\&+\ell_{i,N}(x_{i,N|k+1} - \bar{x}_i^*, \bar{r}_i^*) + \ell_{i,c}^\mu(r_{i,k},g_{i,k},\bar{r}_i^*), 
\end{align}
where $x_{i,N|k+1}$ denotes the terminal state at time $k+1$, given by $x_{i,N|k+1} = f_i(x_{i,N|k}^*, \kappa_i(x_{i,N|k}^*, \bar{r}_i^*))$.
Using the definition of $\ell_{i,N}(\cdot)$ from Assumption \ref{ass:terminal} and positive definiteness of the stage cost function from Assumption \ref{ass:cost}, we obtain the following inequality:
\begin{align}
\label{eq:deltaW}
    \Delta W_i(x_{i,k}, r_{i,k}, g_{i,k}) &\leq -\ell_{i}(x_{i,k} - \bar{x}_i^*, u_{i,k} - \bar{u}_i^*)\nonumber
    \\ &\leq\alpha_l(\norm{x_{i,k} - \bar{x}_i^*}).
\end{align}
Then, since $W_i(x_{i,k}, r_{i,k}, g_{i,k})$ is a Lyapunov function, we can guarantee the closed-loop stability.
To demonstrate the convergence of the robot positions to the optimal Voronoi centroids, i.e.,  $r_{i,k}$, we first show that at each time $k$, the proposed controller leads the robot states to asymptotically converge to $\bar{x}_i^*$. This is evident from the inequality \eqref{eq:deltaW} as follows:
\begin{equation*}
    W_i(x_{i,k+1}, r_{i,k}, g_{i,k}) - W_i(x_{i,k}, r_{i,k}, g_{i,k}) \leq\alpha_l(\norm{x_{i,k} - \bar{x}_i^*}),
\end{equation*}
leads to the following result,
\begin{equation}
    \lim_{k\rightarrow\infty}\norm{x_{i,k} - \bar{x}_i^*} = 0.
\end{equation}
Next, using the results from Lemma \ref{lem:stab_eq}, we have that if $\norm{x_{i,k} - \bar{x}_i^*}=0$, then $W_i(x_{i,k}, r_{i,k}, g_{i,k}) =0$. Therefore, since $\lim_{k\rightarrow\infty}\norm{x_{i,k} - \bar{x}_i^*} = 0$, we obtain
\begin{equation}
    \lim_{k\rightarrow\infty}W_i(x_{i,k}, r_{i,k}, g_{i,k}) = W_{i,\infty} = 0.
\end{equation}
Using the positive definiteness of $W_i(x_{i,k}, r_{i,k}, g_{i,k})$ derived earlier, 
\begin{equation*}
    \lim_{k\rightarrow\infty}\alpha_W(\norm{x_{i,k} - \bar{x}_i^\dagger}) \leq \lim_{k\rightarrow\infty}W_i(x_{i,k}, r_{i,k}, g_{i,k}) = 0.
\end{equation*}
Thus, we get $ \lim_{k\rightarrow\infty}\norm{x_{i,k} - \bar{x}_i^\dagger}= 0$, which proves the result.
\section{Calculation of Terminal Ingredients}
\label{app:terminal}
To ensure the convergence of the proposed nonlinear tracking MPC, we consider an LQR based design \cite{carron2020model,chen1998quasi} for terminal ingredients, i.e., a terminal control law and an invariant set for centroid tracking.
\subsection{Terminal Control Law}
The terminal control law is derived for the linearized system, where the robot dynamics is linearized at the steady-state state-input pair $(\bar{x}_i, \bar{u}_i)$ corresponding to the centroid reference position $\bar{r}_i$. As a result, we get the following state and input matrices.

\begin{equation}
\label{eq:linearized}
    A_{i} = \frac{\partial f_i(x_{i}, u_{i})}{\partial x_i}\bigg|_{(\bar{x}_i, \bar{u}_i)}, \quad B_{i} = \frac{\partial f_i(x_i, u_i)}{\partial u_i}\bigg|_{(\bar{x}_i, \bar{u}_i)}
\end{equation}
The terminal control law $\kappa_i(x_i, \bar{r}_i) = K_i(x_i - \bar{x}_i)$, where $K_i$ is computed using LQR controller design for the linearized robot dynamics \eqref{eq:linearized}. The gain $K_i$ is designed so that the nonlinear robot dynamics is asymptotically stable in the neighborhood of $(\bar{x}_i, \bar{u}_i)$.

\subsection{Terminal Invariant Set for Centroid Tracking}
The following lemma \cite{chen1998quasi} provides a region of attraction, i.e., characterization and design for the terminal invariant set for the robot dynamics controlled by a local linear state-feedback control.
\begin{lemma}
Suppose the closed-loop system $x_{i,k+1} = f_i(x_{i,k}, \kappa_i(x_{i,k}, \bar{r}_i))$ is stabilizable at $(\bar{x}_i, \bar{u}_i)$, and let the linearized closed-loop dynamics under the stabilizing controller $u_{i,k} = \kappa_i(x_{i,k}, \bar{r}_i)$ be defined as $A_{i,K} = A_i+BK_i$. Then,
\begin{enumerate}
    \item There exists a unique positive definite and symmetric solution $P_i$, which satisfies
    \begin{equation}
        \biggl(\frac{A_{i,K}}{\sqrt{1-c_i}}\biggr)^\top P_i\biggl(\frac{A_{i,K}}{\sqrt{1-c_i}}\biggr) - P_i = -Q_i^*
    \end{equation}
    where $Q_i^* = Q + K_i^\top R_i K_i$ is positive definite and symmetric, and $c_i$ satisfies 
    \begin{equation*}
        c_i < 1 - |\lambda_{\text{max}}(A_{i,K})|^2.
    \end{equation*}
    \item The following set $\Gamma_i(\zeta_i)$ is an invariant set for some $\zeta_i\in (0, \infty)$:
    \begin{multline*}
        \Gamma_i(\zeta_i) = \{ (x_{i,k},\bar{r}_i) \in \mathbb{R}^{n_x}\times \mathbb{R}^{d} \ | \\ (x_{i,k} - \bar{x}_i)^\top P_i (x_{i,k} - \bar{x}_i) \leq \alpha_i, \  \bar{r}_i = C_i\bar{x}_i\},
    \end{multline*}
    such that 
    \begin{enumerate}
        \item The controller $u_{i,k} = \kappa_i(x_{i,k}, \bar{r}_i)$ for all $(x_{i,k}, \bar{r}_i) \in \Gamma_i(\zeta_i)$ satisfies the state and input constraints  $(x_{i,k}, u_{i,k}) \in \mathcal{Z}_i$.
        \item The set $\Gamma_i(\zeta_i)$ represents an invariant set for the nonlinear system \eqref{eq:agent_dynamics} controlled by the local control law $u_{i,k} = \kappa_i(x_{i,k}, \bar{r}_i)$.
        \item For any $(x_{i,k}, \bar{r}_i) \in \Gamma_i(\zeta_i)$ with inputs obtained from the local control law $u_{i,k}=\kappa_i(x_{i,k}, \bar{r}_i)$, the infinite horizon cost starting at time $\tilde{k}>0$  
        \begin{equation*}
            J_i^\infty(\tilde{x}_{i}, u_{i,k}) = \sum_{k=\tilde{k}}^\infty x_{i,k}^\top Q_i x_{i,k} + u_{i,k}^\top R_i u_{i,k}
        \end{equation*}
        where $Q_i>0$ and $R_i>0$ represent positive definite matrices. The above cost function starting from $x_{i,\tilde{k}}= \tilde{x}_i$ subject to the nonlinear robot dynamics \eqref{ass:dynamics} is bounded. The bound is given by:
        \begin{equation*}
            J_i^\infty(\tilde{x}_{i}, u_{i,k}) \leq \tilde{x}_{i}^\top P_i \tilde{x}_{i}.
        \end{equation*}
    \end{enumerate}
\end{enumerate}
\end{lemma}
\begin{proof}
The proof is an immediate consequence of the result \cite[Lem. 4]{carron2020model}.   
\end{proof}
\bibliographystyle{ieeetr}
\bibliography{IEEEabrv,root_color} 
\end{document}